\pdfoutput=1

\documentclass[11pt]{article}

\usepackage[]{emnlp2021}

\usepackage{times}
\usepackage{latexsym}

\usepackage{comment}
\usepackage{hyperref}
\usepackage{url}
\usepackage{array,booktabs}
\usepackage{multirow}
\usepackage{adjustbox}
\usepackage{tabularx}
\usepackage{subfig}
\usepackage{amsmath}
\usepackage{amsfonts}
\DeclareMathOperator*{\argmax}{arg\,max}

\usepackage{bbm}
\usepackage[T1]{fontenc}

\usepackage[utf8]{inputenc}

\usepackage{microtype}

\title{Model Selection for Cross-Lingual Transfer}

\author{Yang Chen \textmd{and} Alan Ritter \\
  School of Interactive Computing\\
  Georgia Institute of Technology \\
  \texttt{\{yang.chen, alan.ritter\}@cc.gatech.edu} \\
}

\date{}

\begin{document}
\maketitle
\begin{abstract}
Transformers that are pre-trained on multilingual corpora, such as, mBERT and XLM-RoBERTa, have achieved impressive cross-lingual transfer capabilities.  In the zero-shot transfer setting, only English training data is used, and the fine-tuned model is evaluated on another target language.  
While this works surprisingly well, substantial variance has been observed in target language performance between different fine-tuning runs, and in the zero-shot setup, no target-language development data is available to select among multiple fine-tuned models.
Prior work has relied on English dev data to select among models that are fine-tuned with different learning rates, number of steps and other hyperparameters, often resulting in suboptimal choices.  In this paper, we show that it is possible to select consistently better models when small amounts of annotated data are available in auxiliary pivot languages.
We propose a machine learning approach to model selection that uses the fine-tuned model's own internal representations to predict its cross-lingual capabilities.  In extensive experiments we find that this method consistently selects better models than English validation data across twenty five languages (including eight low-resource languages), and often achieves results that are comparable to model selection using target language development data.\footnote{Our code and data is available at:~\url{ https://github.com/edchengg/model_selection}}
\end{abstract}

\section{Introduction}
Pre-trained Transformers \citep{vaswani2017attention,devlin2019bert} have achieved state-of-the-art results on a range of NLP tasks, often approaching human inter-rater agreement \citep{joshi2020spanbert}.  These models have also been demonstrated to learn effective cross-lingual representations, even without access to parallel text or bilingual lexicons \citep{wu2019beto,pires2019multilingual}.  

In the zero-shot transfer learning, training and development data are only assumed in a high resource source language (e.g. English), and performance is evaluated on another target language.  Because no target language annotations are assumed, source language data is typically used to select among models that are fine-tuned with different hyperparameters and random seeds.  However, recent work has shown that English dev accuracy does not always correlate well with target language performance \citep{keung2020evaluation}.  

In this paper, we propose an alternative strategy for model selection in zero-shot transfer.  Our approach, dubbed Learned Model Selection (LMS), learns a function that scores the compatibility between a fine-tuned multilingual Transformer, and a target language. The compatibility score is calculated based on features of the multilingual model's learned representations.  
This is done by aggregating representations over an unlabeled target language text corpus after fine-tuning on source language data.  We show that these model-specific features effectively capture information about how the cross-lingual representations will transfer.
We also make use of language embeddings from the {\tt lang2vec} package \citep{malaviya2017learning},\footnote{\url{https://github.com/antonisa/lang2vec}} which have been shown to encode typological information that may help inform how a multilingual model will transfer to a particular target.  These model and language features are combined in a bilinear layer to compute a ranking on the fine-tuned models.
Parameters of the ranking function are optimized to minimize a pairwise loss on a set of held-out models, using one or more auxiliary pivot languages.
Our method assumes training data in English, in addition to small amounts of auxiliary language data.
This corresponds to a scenario where the multilingual model needs to be quickly adapted to a new language.
LMS does not rely on any annotated data in the target language, yet it is effective in learning to predict how well fine-tuned representations will transfer.

\begin{figure}[t!]
  \centering
  \subfloat[English Development Selection]{\includegraphics[width=0.5\textwidth]{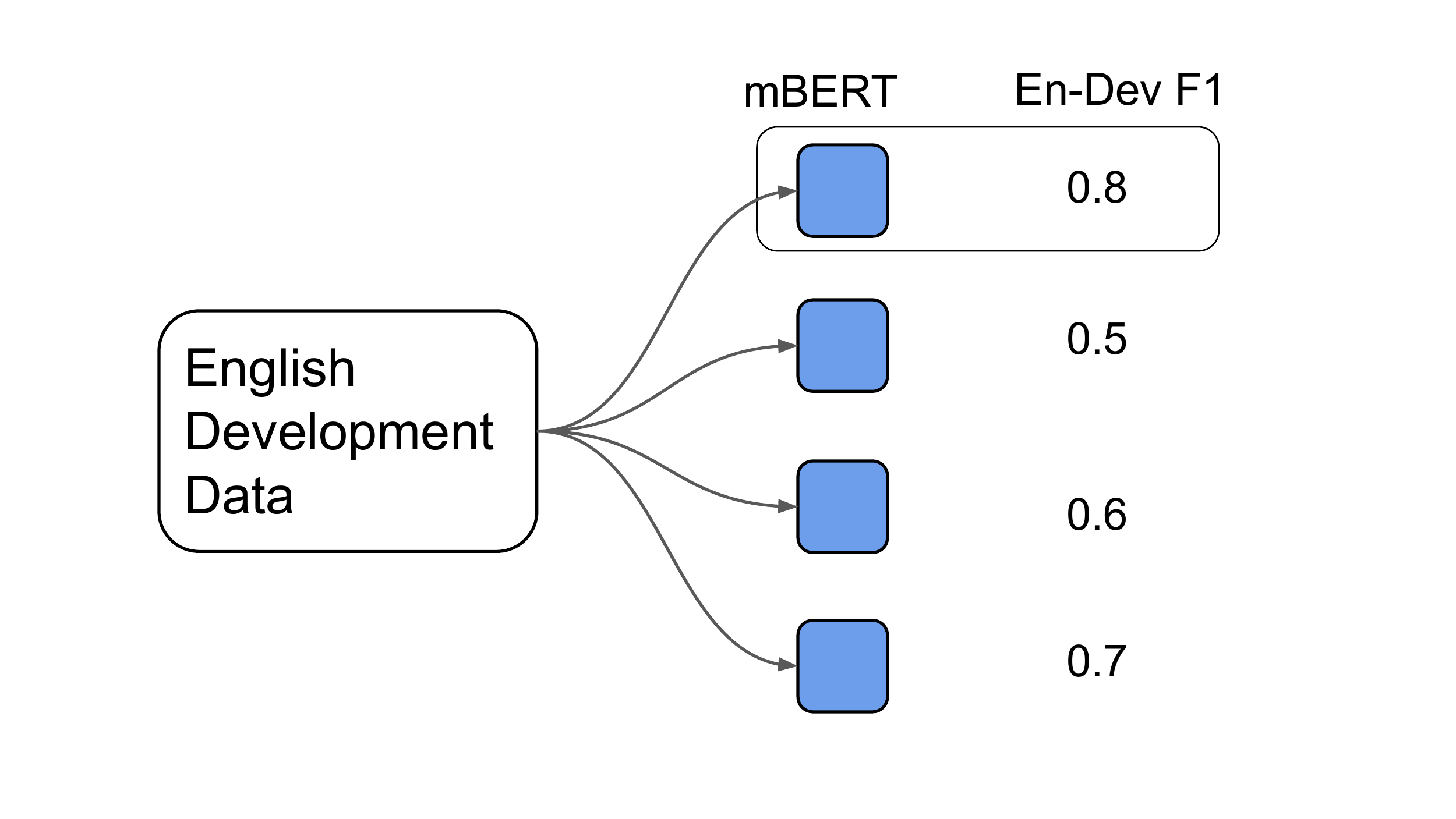}\label{fig:endev}}
  \hfill
  \subfloat[Learned Model Selection (LMS)]{\includegraphics[width=0.5\textwidth]{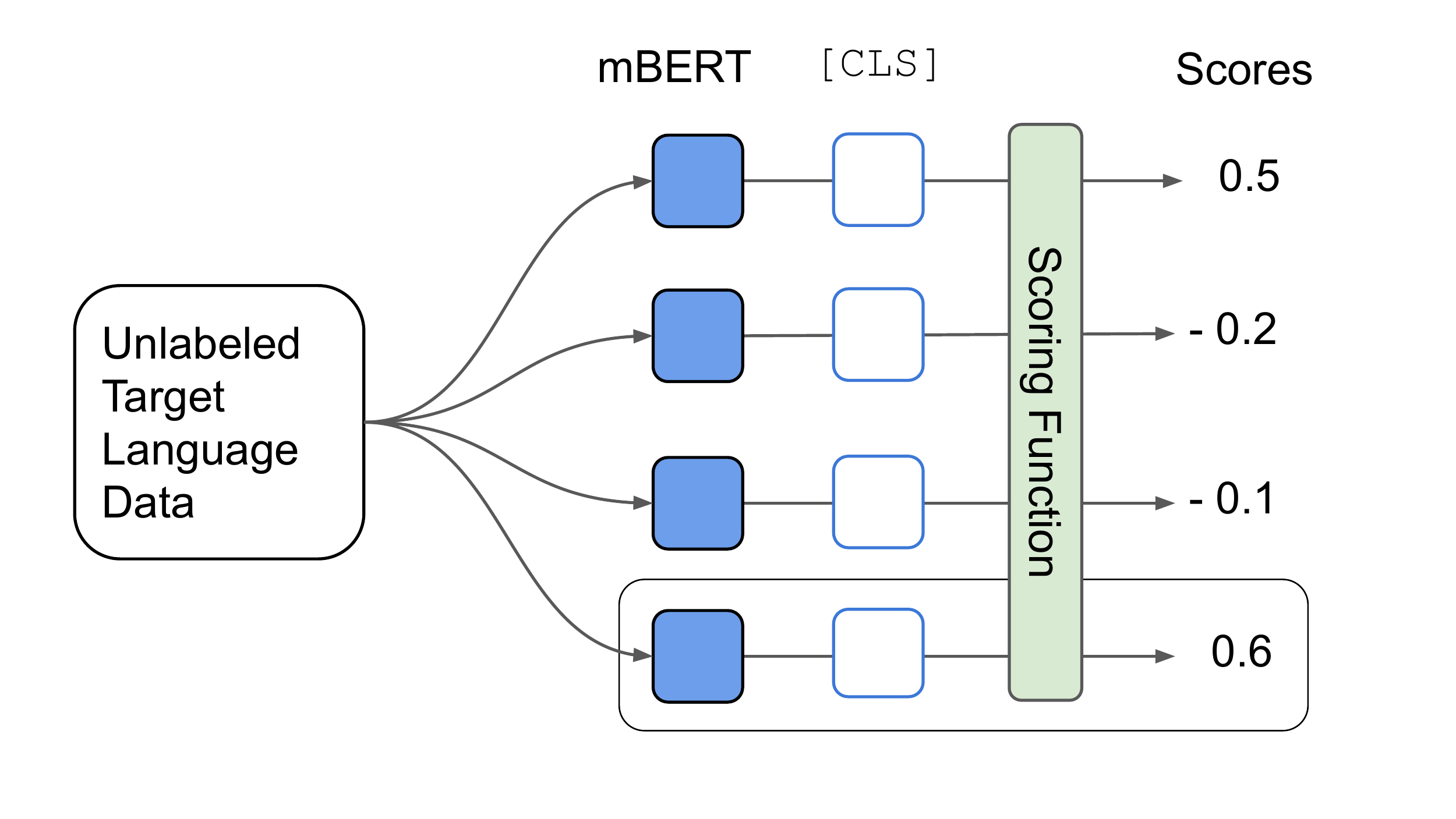}\label{fig:lms}}
  \caption{An illustration of our approach to select the best model for zero-shot cross-lingual transfer. (a) Prior work selects the best model using source language development data. (b) LMS: A learned function scores fine-tuned models based on their hidden layer representations when encoding unlabeled target language data.}
  \label{fig:endev}
\end{figure}

In experiments on twenty five languages, LMS consistently selects models with better target-language performance than those chosen using English dev data.  Furthermore, our proposed approach improves performance on low-resource languages such as Quechua, Maori and Turkmen that are not included in the pretraining corpus (\S \ref{sec:low_resource}). 


\section{Background: Cross-Lingual Transfer Learning}

The zero-shot setting considered in this paper works as follows.  A Transformer model is first pre-trained using a standard masked language model objective.  The only difference from the monolingual approach to contextual word representations \citep{peters2018deep,devlin2019bert} is the pre-training corpus, which contains text written in multiple languages; for example, mBERT is trained on Wikipedia data from 104 languages.  After pre-training, the resulting network encodes language-independent representations that support surprisingly effective cross-lingual transfer, simply by fine-tuning with English data.  For example, after fine-tuning mBERT using the English portion of the CoNLL Named Entity Recognition dataset, the resulting model can perform inference directly on Spanish text, achieving an F$_1$ score around 75, and outperforming prior work using cross-lingual word embeddings \citep{xie2018neural,mikolov2013emb}.  A challenge, however, is the relatively high variance across multiple training runs.  Although mean F$_1$ on Spanish is 75, the performance of 60
fine-tuned models with different learning rates and random seeds ranges from around 70 F$_1$ to 78.  In zero-shot learning, no validation/development data is available in the target language, motivating the need for a machine learning approach to model selection.  

\section{Ranking Model Compatibility with a Target Language}
\label{methods}
Given a set of multilingual BERT-based models, $M=m_1, m_2,..., m_n$ that are fine-tuned on an English training set using different hyperparameters and random seeds, our goal is to select the model that performs best on a specific target language, $l_\text{target}$.  
Our approach (LMS) learns to rank a set of models based on two sources of information: (1) the models' own internal representations, and (2) {\tt lang2vec} representations of the target language \citep{malaviya2017learning}.  

We adopt a pairwise approach to learning to rank \citep{burges2005learning,koppel2019pairwise}. The learned ranking is computed using a scoring function, $s(m,l) = f(g_{\text{mBERT}}(m), g_{{\tt lang2vec}}(l))$, where $g_{\text{mBERT}}(m)$ is a feature vector for model $m$, which is computed from the model's own hidden state representations, and $g_{{\tt lang2vec}}(l)$ is the {\tt lang2vec} representation of language $l$.
The model and language features are each passed through a feed-forward neural network and then combined using a bilinear layer to calculate a final score as follows:

\small
\begin{eqnarray*}
s(m,l) & = & f(g_{\text{mBERT}}(m), g_{{\tt lang2vec}}(l)) \\
& = & \text{\sc ffnn}(g_{\text{mBERT}}(m))^T W_{\text{bi}} \text{\sc ffnn}(g_{{\tt lang2vec}}(l))
\end{eqnarray*}
\normalsize

Using the above score, we can represent the probability that model $m_i$ performs better than $m_j$ on language $l$:
\[
P(m_i \triangleright_l m_j) = \sigma(s(m_i,l) - s(m_j,l))
\]
where $\sigma(\cdot)$ is the sigmoid function.  To tune the parameters of the scoring function, which include the feed-forward and bilinear layers, we minimize cross-entropy loss:
\begin{equation}
\label{eq:cross_entropy}
C = \sum_{l \in L \setminus \{l_\text{target}\}} \sum_{m_i \in M} \sum_{m_j \in M} -C_{m_i,m_j,l}
\end{equation}
\noindent
where
\begin{eqnarray*}
C_{m_i,m_j,l} & = & \mathbbm{1}[m_i \triangleright_l m_j] \log P(m_i \triangleright_l m_j) \\
& & + \mathbbm{1}[m_j \triangleright_l m_i] \log P(m_j \triangleright_l m_i)
\end{eqnarray*}
Here $\mathbbm{1}[m_j \triangleright_l m_i]$ is an indicator function that has the value 1 if $m_j$ outperforms $m_i$, as evaluated using labeled development data in language $l$.  

The first sum in Equation \ref{eq:cross_entropy} ranges over all languages where development data is available (this excludes the target language).  After tuning parameters to minimize cross-entropy loss on these languages, models are ranked based on their scores for the target language, and the highest scoring model, $\hat{m} = \argmax_m s(m,l_\text{target})$, is selected.

\section{Tasks and Datasets}

We perform model selection experiments on five well-studied NLP tasks in the zero-shot transfer setting: part-of-speech (POS) tagging, 
question answering (QA),
relation extraction (RE), event-argument role labeling (ARL), and named entity recognition (NER).
In total, we cover 25 target languages including 8 low-resource languages in our experiments following prior work (shown in Table~\ref{table:tasks}). We adopt the best performing model from \citet{soares2019re}, [{\sc Entity Markers - Entity Start}], for RE and ARL. For other tasks, we use established task-specific layers and evaluation protocols, following the references in Table \ref{table:tasks}.
Labeled training data for each task is assumed in English and trained models are evaluated on each target language. 




\begin{table*}[ht!]
    \centering
    \begin{tabular}{llll}
        \toprule
                Task& Dataset & References & Target Languages \\
         \midrule
         \multirow{2}{*}{POS} & \multirow{2}{*}{UD~\citep{ud14}} & \multirow{2}{*}{\citet{wu2019beto}} & \textit{ar,bg,da,de,es,fa,hu,it}\\
         & & & \textit{nl,pt,ro,sk,sl,sv,vi,zh}\\
         QA& MLQA~\citep{lewis2019mlqa} & \citet{lewis2019mlqa} & \textit{ar, de, es, hi, vi, zh}\\ 
         RE& ACE05~\citep{walker2006ace} & \citet{sub2019rearl}& \textit{ar, zh}\\
         ARL& ACE05~\citep{walker2006ace} & \citet{sub2019rearl} & \textit{ar, zh}\\
         NER & CoNLL~\citep{sang2002conll} & \citet{wu2019beto} & \textit{de, es, nl, zh}\\
         \midrule
         \multicolumn{4}{c}{\emph{Low-resource languages}}\\
         \midrule
         \multirow{2}{*}{NER} & \multirow{2}{*}{WikiAnn~\citep{pan2017wikiann}} & \multirow{2}{*}{\citet{pfeiffer2020madx,xia2021meta}} & \textit{cdo, gn, ilo, mhr}\\
         & & & \textit{mi, tk, qu, xmf}\\
         \bottomrule
    \end{tabular}
\caption{25 target languages and five tasks used in our experiments. English is used as the source language. ar: Arabic, bg: Bulgarian, da: Danish, de: German, es: Spanish, fa: Persian, hi: Hindi, hu: Hungarian, it: Italian, nl: Dutch, pt: Portuguese, ro: Romanian, sk: Slovak, sl: Slovene, sv: Swedish, vi: Vietnamese, zh: Chinese. Low-resource languages information can be found in Table~\ref{table:low_resource}.}
\label{table:tasks}
\end{table*}

\subsection{Low-resource Languages}
To evaluate LMS on truly low-resource languages, we use the 8 target languages (summarized in Table~\ref{table:low_resource}) following~\citep{pfeiffer2020madx, xia2021meta} which uses the WikiAnn NER dataset~\citep{pan2017wikiann}. These languages are considered low-resource because: 1) the Wikipedia size ranges from 4k to 22k; 2) they are not covered by pre-trained multilingual models (i.e., by mBERT and XLM-RoBERTa).
The train, development, and test partition of \citet{rahimi2019ner} is used following the XTREME benchmark's NER setup~\citep{hu2020xtreme}. 
The related language used for the Pivot-Dev baseline is chosen following \citet{xia2021meta}, which is based on LangRank~\citep{lin2019choosing}.

\begin{table}[ht]
    \centering
    \scriptsize
    
    \begin{tabularx}{\columnwidth}{llllr}
    \toprule
    \multirow{2}{*}{Language}  & \multirow{2}{*}{Code}& Language & Related &$\#$Wiki\\
                  &  & Family   & Language &articles \\
    \midrule
    Min Dong & cdo & Sino-Tibetan & Chinese (zh) & 15k\\
    Guarani & gn & Tupian & Spanish (es) & 4k\\
    Ilocano & ilo & Austronesian & Indonesian (id) & 14k\\
    Meadow Mari & mhr & Uralic & Russian (ru) & 10k\\
    Maori & mi & Austronesian & Indonesian (id) & 7k\\
    Turkmen & tk & Turkic & Turkish (tr) & 6k\\
    Quechua & qu & Quechua & Spanish (es) & 22k\\
    Mingrelian & xmf & Kartvelian & Georgian (ka) & 13k\\
    \bottomrule
    \end{tabularx}
\caption{Low-resource target languages used in the WikiAnn NER task. Related languages, used in the Pivot-Dev baseline, are selected following~\citep{xia2021meta}.}
\label{table:low_resource}
\end{table}

\section{Experimental Design}
\label{exp:design}
For a multilingual NLP task with $n$ languages $L:\{l_1,...,l_n\}$, our goal is to select the model that performs best on a new target language, $l_{\text{target}} \not\in L$. We assume the available resources are English training and development data, in addition to a small development set for each of the pivot languages, $L$.
First, a set of mBERT models, $M$, are fine-tuned on an English training set using different hyperparameters and random seeds and shuffled into meta-train/dev/test sets. We then evaluate each model, $m_i$, on the pivot languages' dev sets to calculate a set of gold rankings, $\triangleright_l$, that are used in the cross-entropy loss (Equation \ref{eq:cross_entropy}). Model-specific features are extracted from the fine-tuned mBERTs, by feeding unlabeled pivot language text as input.

\noindent
{\bf Development and Evaluation}
mBERT models in the meta-dev set are used to experiment with different model and language features.
Evaluation is performed using models in the meta-test set. We use the leave-one-language-out setup for each task during evaluation.
For each target language, we rank models using the learned scoring function, select the highest scoring model, and report results in Table~\ref{table:all}.

\subsection{Baselines and Oracles} 
En-Dev is our main baseline following standard practice for model selection in zero-shot transfer learning~\citep{wu2019beto,pires2019multilingual}. 
Because our approach assumes additional development data in  auxiliary languages, we also include a baseline that uses pivot-language language dev data.\footnote{The auxiliary language with highest similarity to the target language, as measured using cosine similarity between {\tt lang2vec} embeddings, is used in this baseline.}
In addition, we compare against an oracle that selects models using 100 annotated sentences from the target language dev set to examine how our approach compares with the more costly alternative of annotating small amounts of target language data.  Finally, we include an oracle that simply picks the best model using the full target language development set (All-Target).  All baselines and oracles are summarized below:

\begin{itemize}
    \item En-Dev (baseline): chooses the fine-tuned mBERT with best performance on the English dev set.
    \item Pivot-Dev (baseline): chooses the fine-tuned mBERT with best performance on development data in the most similar pivot language (similarity to the target language is measured using {\tt lang2vec} embeddings).
    \item 100-Target (oracle): chooses the fine-tuned mBERT with best performance on 100 labeled target language instances.
    \item All-Target (oracle): chooses the fine-tuned mBERT using the full target language dev set.
\end{itemize}

\subsection{Hyperparameters and Other Settings}
\label{exp:hyper}
To train the scoring function, $s(\cdot)$, we use Adam~\citep{kingma2014adam}, and select the batch size among \{16, 32, 64, 128\}, learning rate $\lambda$ among $\{1\times10^{-4}, 5\times10^{-5}, 1\times10^{-5}, 5\times10^{-6}, 1\times10^{-6}\}$, and train for \{3\} epochs. The scoring function, $s(\cdot)$, contains a 2-layer $\text{\sc ffnn}$ with 1024 hidden units and ReLU activation \citep{glorot2011relu}. 
The base cased mBERT has 179M parameters and a vocabulary of around 120k wordpieces.  Both the pre-trained Transformer layers and task-specific layers are fine-tuned using Adam, with $\beta_1=$ 0.9, $\beta_2=$ 0.999, and an L2 weight decay of 0.01.  Model candidates are fine-tuned with varying learning rates and number of epochs with the following settings: learning rate $ \in \{3\times10^{-5}, 5\times10^{-5}, 7\times10^{-5}\}$; number of epochs $\in \{3,4,5,6\}$; batch size $ \in \{32\}$; random seeds $\in \{0,1,...,239\}$. 240 mBERT models with different random seeds are fine-tuned with 12 different hyperparameter settings (20 random seeds for each set of hyperparameters), and then split into meta-train/dev/test sets (120/60/60). All models are trained on an RTX 2080 Ti.

\section{Evaluation}
\label{exp:res}
Below we report model selection results on mBERTs in the meta-test set for each of the five tasks.

\begin{table*}[ht!]
\centering
\small
\begin{tabular}{ll|c|cll|ccl}
\toprule
    Task  & Lang & Ref  & En-Dev & Pivot-Dev & LMS            & 100-Target & All-Target  & \# All-Target \\ 
\midrule
\multirow{15}{*}{POS (Acc)}
&ar & - & 49.7 & \underline{50.3} (de)$^{**}$ & \textbf{51.6}$^{**}$ & 50.6 & 52.7 & 786 \\
&de & 89.8 & 89.3 & \underline{88.7} (nl) & \textbf{89.8}$^{**}$ & 89.4 & 89.9 & 799 \\
&es & 85.2 & 84.8 & \underline{85.3} (nl)$^{**}$ & \textbf{85.6}$^{**}$ & 84.8 & 85.1 & 1552 \\
&nl & 75.9 & 75.7 & \textbf{75.9} (de)$^{**}$ & \textbf{75.9}$^{**}$ & 75.5 & 76.0 & 349 \\
&zh & - & \underline{66.9} & \underline{66.9} (de)$^{**}$ & \textbf{68.0}$^{**}$ & 67.3 & 68.8 & 500 \\
&bg & 87.4 & \underline{87.1} & 87.0 (es) & \textbf{87.9}$^{**}$ & 87.9 & 87.9 & 1115 \\
&da & 88.3 & 88.6 & \underline{88.8} (nl)$^{**}$ & \textbf{88.9}$^{**}$ & 88.6 & 89.2 & 322 \\
&fa & 72.8 &\underline{71.6} & \underline{71.6} (es) & \textbf{73.6}$^{**}$ & 73.6 & 73.7 & 599 \\
&hu & 83.2 & \underline{82.5} & 82.0 (de) & \textbf{83.3}$^{**}$ & 83.3 & 83.1 & 179 \\
&it & 84.7 & 84.5 & \underline{84.9} (es)$^{**}$ & \textbf{85.2}$^{**}$ & 85.4 & 85.8 & 489 \\
&pt & 82.1 & \underline{81.8} & \underline{81.8} (es) & \textbf{82.2}$^{**}$ & 81.8 & 82.2 & 271 \\
&ro & 84.7 & 83.8 & \underline{84.2} (es)$^{**}$ & \textbf{84.7}$^{**}$ & 84.4 & 85.4 & 1191 \\
&sk & 83.6 & \underline{83.7} & 83.6 (es) & \textbf{84.2}$^{**}$ & 83.6 & 84.8 & 1060 \\
&sl & 84.2 & \underline{84.5} & 83.6 (es) & \textbf{85.2}$^{**}$ & 83.8 & 85.5 & 735 \\
&sv & 91.3 & 91.4 & \textbf{91.8} (nl)$^{**}$ & \underline{91.7}$^{**}$ & 91.3 & 91.8 & 504 \\
\cmidrule{2-9} 
     & AVG En-Dev $\Delta$ & - & 0.0 & 0.0 & \textbf{0.9} & 0.3 & 1.0 & -\\
\midrule
\multirow{6}{*}{QA (F$_1$)}
&ar & 45.7 & 47.7 & \textbf{49.4} (de)$^{**}$ & \underline{49.3}$^{**}$ & 49.4 & 49.4 & 517 \\
&de & 57.9 & 55.3 & \underline{55.8} (ar)$^{**}$ & \textbf{55.9}$^{**}$ & 57.1 & 55.8 & 512 \\
&es & 64.3 & \underline{64.9} & 64.7 (ar) & \textbf{65.0} & 64.5 & 65.1 & 500 \\
&zh & 57.5 & \underline{58.0} & \underline{58.0} (de) & \textbf{58.1} & 58.1 & 58.4 & 504 \\
&hi & 43.8 & 39.1 & \underline{42.1} (es)$^{**}$ & \textbf{42.4}$^{**}$& 38.8& 42.9 & 507\\
&vi & 57.1 & \underline{57.3} & 56.9 (ar) & \textbf{58.2}$^{**}$ & 59.1 & 58.1 & 511\\
\cmidrule{2-9} 
     & AVG En-Dev $\Delta$ & - & 0.0 & 0.8 & \textbf{1.1} & 0.8 & 1.2 & -\\
\midrule
\multirow{2}{*}{RE (F$_1$)}
&ar & 39.4 & \underline{36.1} & 35.3 (zh) & \textbf{39.5}$^{**}$ & 34.7 & 41.9 & 4482 \\
&zh & 32.7 & \underline{67.7} &  67.4 (ar) & \textbf{70.8}$^{**}$ & 68.2 & 69.1 & 7096 \\
\cmidrule{2-9} 
     & AVG En-Dev $\Delta$ & - & 0.0 & -0.6 & \textbf{3.3} & -0.5 & 3.6 & -\\
\midrule
\multirow{2}{*}{ARL (F$_1$)}
&ar & 16.5 & 44.1 & \textbf{48.1} (zh)$^{**}$ & \underline{47.1}$^{**}$ & 44.1 & 47.2 & 1221 \\
&zh & 23.5 & 61.0 & \underline{61.3} (ar) & \textbf{62.1}$^{**}$ & 62.5 & 63.8 & 2226 \\
\cmidrule{2-9} 
     & AVG En-Dev $\Delta$ & - & 0.0 & \textbf{2.2} & 2.1 & 0.8 & 3.0 & -\\
\midrule
\multirow{4}{*}{NER$_{\text{CoNLL}}$ (\text{F}$_1$)}
     &de & 69.6 & 69.9 & \underline{70.7} (nl)$^{**}$ & \textbf{71.0}$^{**}$ & 66.7 & 72.1  & 2867 \\ 
     &es & 75.0 & \underline{74.6} & 73.1 (nl) & \textbf{75.7}$^{**}$ & 75.7 & 75.7 & 1915 \\
     &nl & 77.6 & 78.7 & \textbf{79.3} (de)$^{**}$ & \underline{78.9}$^{**}$ & 78.7 & 80.3 & 2895 \\
     &zh & 51.9 & \underline{54.9} & 53.0 (de) & \textbf{55.1} & 55.4 & 56.9  & 4499 \\
     \cmidrule{2-9} 
     & AVG En-Dev $\Delta$ & - & 0.0 & -0.5 & \textbf{1.1} & -1.1 & 2.1 & -\\
\midrule
\multicolumn{9}{c}{\emph{Low-resource languages}}\\
\midrule
\multirow{8}{*}{NER$_{\text{WikiAnn}}$ (\text{F}$_1$)}  & cdo & 14.2 & 11.0 & \underline{12.4} (zh) & \textbf{19.4}$^{**}$ & 19.4 & 19.4 & 100\\
& gn & 45.4 & 45.0 & \underline{47.1} (es) & \textbf{49.0}$^{**}$ & 46.2 & 46.2 &  100\\
& ilo& 63.5 & \textbf{61.4} & 59.2 (id) & \underline{61.1} & 66.1 & 66.1 &  100\\
& mhr& 46.0 & \underline{45.3} & 41.1 (ru) & \textbf{48.6}$^{**}$ & 48.3 & 48.3 &  100\\
& mi& 21.8 & 28.2 & \underline{33.8} (id)$^{**}$ & \textbf{43.7}$^{**}$ & 55.6 & 55.6 &  100\\
& tk& 47.2 & 51.6 & \textbf{55.5} (tr)$^{**}$ & \underline{54.9}$^{**}$ & 56.5 & 56.5 &  100\\
& qu& 54.9 & \underline{59.8} & \textbf{62.2} (es)$^{**}$ & \underline{59.8} & 63.4 & 63.4 &  100\\
& xmf&31.1 & 34.3 & \underline{36.4} (ka)$^{**}$ & \textbf{37.5}$^{**}$ & 37.8 & 37.8 &  100\\
 \cmidrule{2-9} 
& AVG En-Dev $\Delta$ & - & 0.0 & 1.4 & \textbf{4.7} & 7.1 & 7.1 & -\\
\bottomrule
\end{tabular}
\caption{Model scores selected based on LMS for POS, QA, RE, ARL, and NER. All mBERT models are fine-tuned on English training data. En-Dev / Pivot-Dev / 100-Target / All-Target: model selection based on the highest F1 of English dev set / Pivot language dev set (pivot language in bracket) / 100 target language dev set examples / target language dev set. LMS: model selection based on the highest scores for the target language: $\argmax_m s(m,l_\text{target})$; ``\# All-Target'' is the number of labeled target-language sentences used for model selection in the All-Target oracle. Bold / underlined indicates the best / second best. AVG En-Dev$\Delta$: average differences with En-Dev baseline. Significance compared to the En-Dev is indicated with $^{\ast\ast}(p < 0.05)$ – all tests are computed using the
paired bootstrap procedure~\citep{berg2012sign}.}
\label{table:all}
\end{table*}

\noindent
{\bf POS} Table~\ref{table:all} presents POS accuracies on the test set, using various approaches to model selection for the fifiteen target languages. 
LMS outperforms En-Dev and Pivot-Dev except in the case of Swedish (sv) and Dutch (nl). Interestingly, model selection for Italian with Spanish dev set does not outperform LMS. We use~\citep{wu2019beto} as references for zero-shot cross-lingual transfer with mBERT.

\noindent
{\bf QA} Our method selects a model with higher F1 across all languages compared with En-Dev, although we find that Pivot-Dev performs slightly better on Arabic (ar).
We use~\citep{lewis2019mlqa} as references for zero-shot cross-lingual transfer with mBERT.

\noindent
{\bf ARL and RE} In Table~\ref{table:all}, our method selects models with higher F1 scores compared to En-Dev. It also outperforms 100-Target on Arabic.  We hypothesize this is because 100 target-language examples is not sufficient for effective model selection, as the dataset contains a large proportion of negative examples (no relation).  Also, RE and ARL have large label sizes (18 and 35) so a random sample of 100 instances might not cover every label.  In contrast, the full dev set contains thousands of examples.
We use a Graph Convolutional Network ($\text{GCN}$)~\citep{sub2019rearl} as a reference (see Appendix \ref{appendix_imb_data} for details) and models were selected using the English dev set.

\noindent
{\bf CoNLL NER} As illustrated in Table~\ref{table:all}, our method selects models with a higher F1 score than En-Dev. Besides, it outperforms model selection using small amounts of target-language annotations (100-Target) on Dutch (nl) and German (de) and selects a model that performs as well on Spanish (es). 
On average, LMS achieves 1.6 point increase in F1 score relative to Pivot-Dev.
We use~\citep{wu2019beto} as references for zero-shot cross-lingual transfer with mBERT.

\begin{figure}[!h]
  \centering
  \subfloat{\includegraphics[width=0.5\textwidth]{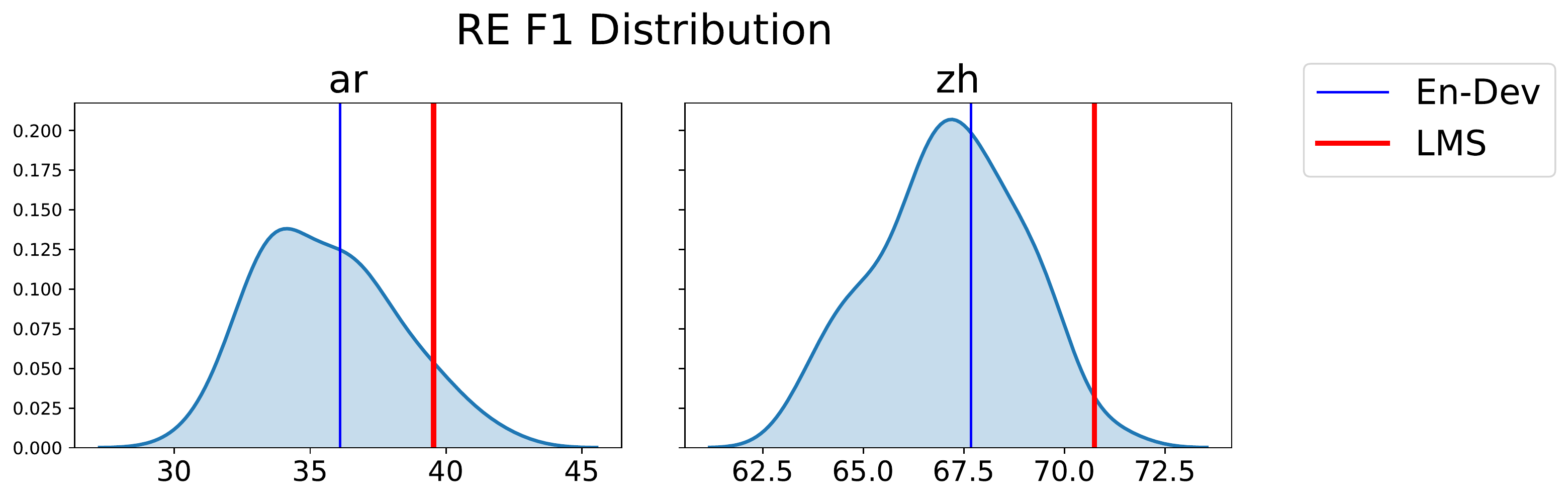}\label{fig:re}}
  \hfill
  \subfloat{\includegraphics[width=0.5\textwidth]{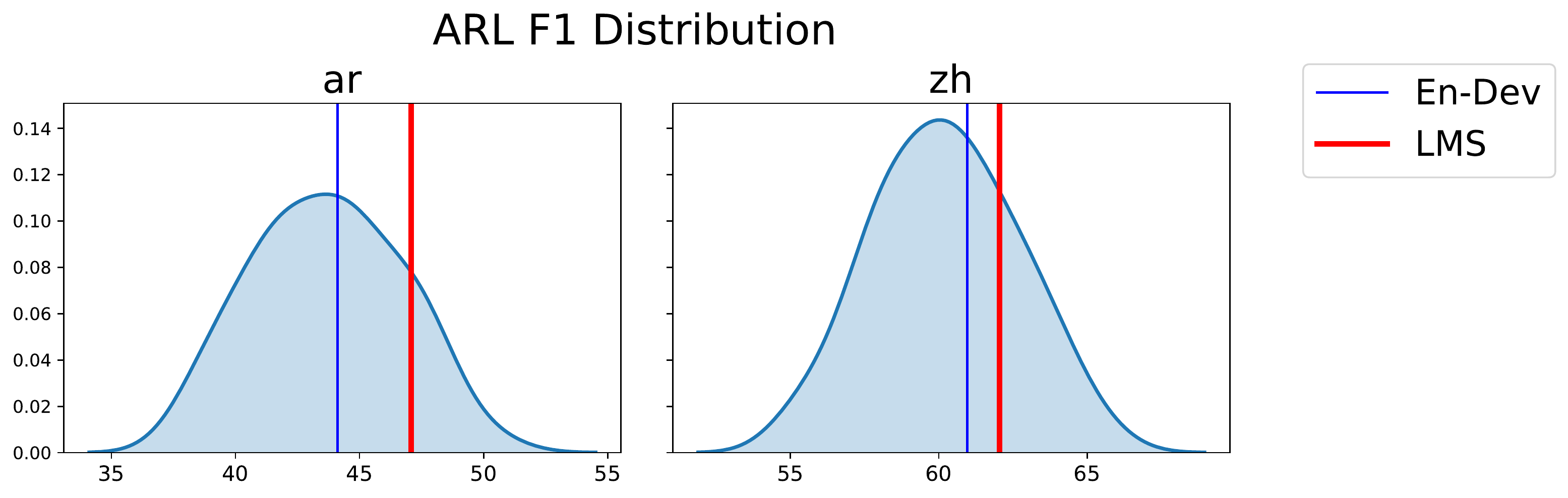}\label{fig:arl}}
  \caption{Model F1 score distributions for RE and ARL. Red line: LMS and blue line: En-Dev. X-axis is F1 score. Selecting models with LMS achieve better results compared to En-Dev.}
  \label{fig:re_arl}
\end{figure}

\noindent
{\bf Model Score Distributions} 
Figure \ref{fig:re_arl} visualizes the En-Dev and LMS results on the test set in the context of the score distributions of the 60 models in the meta-test set, using kernel density estimation. English development data tends to select models that perform only slightly better than average, whereas LMS does significantly better. 

\subsection{Evaluation on low-resource languages}
\label{sec:low_resource}
We present results of low-resource languages NER in the bottom section of Table~\ref{table:all}, where we use 40 pivot languages in the XTREME benchmark~\citep{hu2020xtreme} to train LMS and test on 8 target languages. LMS, outperforms the En-Dev and Pivot-Dev baselines, leading to an average gain of 4.7 and 3.3 F1 respectively. 
Since the setting is targeting truly low-resource languages where \texttt{lang2vec} might not be available, the scoring function thus directly predicts a score based on the representation from unlabeled target language text. We use the mBERT zero-shot cross-lingual transfer results from~\citep{pfeiffer2020madx} as references.

\begin{table*}[ht!]
\centering
\small
\begin{tabular}{ll|c|cll|ccl}
\toprule
    Task  & Lang & Ref  & En-Dev & Pivot-Dev & LMS            & 100-Target & All-Target  & \# All-Target \\ 
\midrule
\multirow{10}{*}{POS (Acc)}
&bg & 87.4 & \underline{90.3} & \underline{90.3} (es) & \textbf{90.8}$^{**}$ & 90.6 & 90.8 & 1115 \\
&da & 88.3 & \textbf{89.4} & 89.2 (nl) & \underline{89.3} & 88.6 & 89.2 & 322 \\
&fa & 72.8 &\underline{80.8} & 79.9 (es) & \textbf{81.1}$^{**}$ & 89.4 & 89.7 & 599 \\
&hu & 83.2 & \textbf{84.2} & \underline{84.1} (de) & 84.0 & 82.1 & 82.1 & 179 \\
&it & 84.7 & \underline{94.0} & \underline{94.0} (es) & \textbf{94.9}$^{**}$ & 95.1 & 95.1 & 489 \\
&pt & 82.1 & \underline{91.1} & \textbf{91.3} (es) & \underline{91.1} & 91.1 & 91.5 & 271 \\
&ro & 84.7 & \underline{88.7} & \underline{88.7} (es) & \textbf{88.8}$^{**}$ & 89.1 & 89.1 & 1191 \\
&sk & 83.6 & \underline{88.3} & \underline{88.3} (es) & \textbf{88.6}$^{**}$ & 89.0 & 89.0 & 1060 \\
&sl & 84.2 & 86.1 & \underline{86.4} (es) & \textbf{86.8}$^{**}$ & 86.8 & 86.8 & 735 \\
&sv & 91.3 & \underline{91.8} & 90.9 (nl) & \textbf{92.2}$^{**}$ & 92.5 & 92.5 & 504 \\
\cmidrule{2-9} 
     & AVG En-Dev $\Delta$ & - & 0.0 & 0.0 & \textbf{0.1} & 0.6 & 0.7 & -\\
\bottomrule
\end{tabular}
\caption{Model scores selected based on LMS for POS. En-Dev / Pivot-Dev / 100-Target / All-Target: model selection based on the highest F1 of English dev set / Pivot language dev set (pivot language in bracket) / 100 target language dev set examples / target language dev set. All mBERT models are fine-tuned on English training data in addition with a small amount of data from pivot languages in $\{ar,de,es,nl,zh\}$. We use English data fine-tuned mBERT results from~\citet{wu2019beto} as references (Ref). Significance compared to the En-Dev is indicated with $^{\ast\ast}(p < 0.05)$ – all tests are computed using the
paired bootstrap procedure~\citep{berg2012sign}.}
\label{table:pos}
\end{table*}

\subsection{Evaluation on multilingual fine-tuned models}
An interesting question is whether fine-tuning on available development data in the auxiliary languages can improve performance.
Since our model assumes access to small amounts of labeled data in a set of pivot languages, we experiment with multilingual fine-tuning and show LMS is still beneficial for selecting among models that are fine-tuned on both English and pivot language data.

Part of speech tagging experiments are presented in Table~\ref{table:pos}, where all mBERT models are fine-tuned on English and the development sets of five pivot languages (ar,de,es,nl,zh).
A single LMS is then trained using English fine-tuning, with gold rankings computed on the pivot languages. Then, we directly apply the English LMS model to do model selection on the multilingual fine-tuned mBERT models.
We find LMS outperforms the En-Dev baseline on seven out of the ten target languages used in our evaluation, with an average gain of 0.1 accuracy. This also demonstrates LMS that is trained on English fine-tuned representations generalizes to multilingual fine-tuning.
We use models that are fine-tuned only on English data, from ~\citet{wu2019beto}, as references, and find that multilingual fine-tuning shows better cross-lingual transfer performance compared to fine-tuning on only English data.

\section{Analysis}
In Section~\ref{exp:res}, we empirically demonstrated that our learned scoring function, $s(\cdot)$, consistently selects better models than the standard approach (En-Dev), and is comparable to small amounts of labeled target language data. Section~\ref{sec:inputfeature} presents additional analysis of our approach, exploring the impact of various modeling choices with $\{ar,de,es,nl,zh\}$. In addition, analysis of generalization beyond mBERT and across tasks capability are present in Appendices~\ref{appendix_xlm} and~\ref{appendix_multi_task}. 

\subsection{Model and Language Features}
\label{sec:inputfeature}
This section explores the impact of different choices for model and language representations for LMS.  Four types of model features and two language embeddings are explored.  We start by delineating possible choices for representations, then describe the details of our experiments, results, and the final choices used in \S \ref{exp:res}.

Four model-specific features are described below. Note {\tt [CLS]} vectors are extracted from mBERT by feeding unlabeled text as input.
\begin{itemize}
   \item {\tt [Eng]}: Averaged {\tt [CLS]} vectors computed over an unlabeled English text corpus are used for both training and testing.\footnote{In our experiments, sentences in the English dev set are used for this purpose (ignoring the labels).}
   \item {\tt[Pivot]}: During training, {\tt [CLS]} vectors are averaged over an unlabeled text corpus in the pivot language.  At test time, {\tt [CLS]} embeddings are averaged over an unlabeled corpus in the target language.  We use the target-language development set (ignoring labels) for this purpose in our experiments.
   \item {\tt [Target]}: {\tt [CLS]} vectors are averaged over a text corpus in the target language (for both training and testing).
   \item Fusion: A linear combination of the above features.  Weights on each representation are learned during training.
\end{itemize}

\noindent
Two types of language embeddings are examined.
\begin{itemize}
    \item {\tt lang2vec}: 512-dimensional vectors learned by a neural network trained for typological prediction~\citep{malaviya2017learning}.
    \item {\tt syntax}: 103-dimensional binary vectors, which capture syntax features from the URIEL knowledge base~\citep{littell2017uriel}.
\end{itemize}

First, we determine the choice of model-specific features by averaging performance across both language embeddings. Table~\ref{table:repr} reports averaged evaluation metrics for each model-specific representation across all target languages with En-Dev as a baseline.

Averaged evaluation metrics across all target languages for each language embedding are reported in Table~\ref{table:lang}. In addition to evaluating the effectiveness of each language embedding, we also experimented with a variant of our scoring function that does not include any language embeddings as input.  Results are reported on mBERT models in the meta-dev set and the target languages' dev sets for all experiments in this section.

\begin{table}[h!]
\scriptsize
\begin{center}
\begin{tabular}{lc|cccc}
\toprule
 \emph{Task} & \emph{En-Dev}& {\tt [Eng]} & {\tt [Pivot]} & {\tt [Target]} & Fusion  \\
 \midrule
 POS & 74.69 & {\bf 75.58} & \underline{75.54} & 75.48 & 75.04\\
QA & 56.31 & 56.49 & \textbf{56.79} & \underline{56.68} & 56.63\\
 RE & 51.81 & 54.92 & {\bf 55.57} & \underline{55.56} & 54.57 \\
 ARL & 50.98 & 51.99 &  \underline{53.74} & 52.31 & \textbf{54.69}\\
 NER & 70.45 & 70.64 & \underline{71.18} & \textbf{71.87} & 70.66 \\
 \midrule
Avg & 60.85 & 61.92 & {\bf 62.60}& \underline{62.38} & 62.32\\
 \midrule
\end{tabular}
\end{center}
\caption{Model-specific feature analysis. We use mBERT models in the meta-dev set for analysis. Each number represents average of scores across all the target languages in a particular task.
  }
\label{table:repr}
\end{table}

\begin{table}[h!]
\scriptsize
\begin{center}
\begin{tabular}{lc|ccc}
\toprule
 \emph{Task} & \emph{En-Dev}& {\tt lang2vec} & {\tt syntax} & None \\
 \midrule
 POS & 74.69 & {\bf 75.72} & 75.36 & 75.20\\
 QA & 56.31 & \textbf{56.81} & 56.77 & 56.49\\
 RE & 51.81 & \textbf{55.92} & 55.22 & 52.53\\
 ARL & 50.98 & 53.60 & \textbf{53.88} & 53.14 \\
 NER & 70.45 & \textbf{71.37} & 70.98 & 70.08\\
 \midrule
 Avg & 60.85 &  \textbf{62.68} & 62.44 & 61.49\\ 
 \midrule
\end{tabular}
\end{center}
\caption{Language embedding analysis across {\tt lang2vec}, {\tt syntax}, and no language embedding. We use  mBERT models in the meta-dev set for analysis. Each number represents average of scores across all the target languages in a particular task.
  }
\label{table:lang}
\end{table}

In Table~\ref{table:repr}, {\tt [PIVOT]} features achieve top-2 performance in all five tasks. {\tt [Eng]} and {\tt [Target]} achieve mixed results, and the fusion of three features does not effectively incorporate the advantages of each representation, except in the case of ARL.  
Table~\ref{table:lang} shows that {\tt lang2vec} outperforms {\tt syntax} for all tasks but ARL and also outperforms our approach when language embeddings are not included. Thus, {\tt lang2vec} and {\tt [PIVOT]} are used for all experiments in Section~\ref{exp:res}.

\section{Related Work}

Recent work has explored hyper-parameter optimization \cite{klein2019meta}, and model selection for a new task. 
{\tt task2vec}~\citep{aless2019task2vec} presents a meta-learning approach to selecting a pre-trained feature extractor from a library for a new visual task. More concretely, {\tt task2vec} represents tasks in a vector space and is capable of predicting task similarities and taxonomic relations. It encodes a new task and selects the best feature extractor trained on the most similar task. Unlike {\tt task2vec}, we select a trained model for a specific task, and we represent a trained model with model-specific features on a target language. 

MAML \citep{finn2017model,rajeswaran2019meta} is another approach to meta-learning, pre-training a single model with a meta-loss to initialize a set of parameters that can be quickly fine-tuned for related tasks.  \citet{nooralahzadeh2020zero} explore the use of MAML in the cross-lingual transfer setting.
MAML is designed to support few-shot learning through better initialization of model parameters and does not address the problem of model selection.  In contrast, our approach improves model selection in the zero-shot cross-lingual transfer setting.

Most relevant to our work,~\citet{xia2020predict} use regression methods to predict a model's performance on an NLP task.  They formulate this as a regression problem based on features of the task (dataset size, average sentence length, etc.), incorporating a discrete feature to represent the choice of model. In contrast, LMS inspects a model's internal representations, thus it is suitable for predicting which out of a set of fine-tuned models will best transfer to a target language.  Also relevant is prior work on learning to select the best language to transfer from \citep{lin2019choosing}.

There is a need for more NLP research on low-resource languages \citep{joshi2020state}.  \citet{lauscher-etal-2020-zero} present a number of challenges in transferring to languages with few resources using pre-trained Transformers.  
Our experiments do cover a set of 8 truly low-resource languages following prior work~\citep{pfeiffer2020madx,xia2021meta} and a fairly diverse set of languages, including Arabic and Chinese.
We believe that there is still a need for more research on multilingual NLP for high-resource languages as well, as this is not a solved problem.  
Finally, we note that there are several other prominent benchmarks for evaluating cross-lingual transfer including XTERME~\citep{hu2020xtreme} and XGLUE~\citep{liang2020xglue}, both of which include some datasets used in this work.

\section{Conclusion}
\label{conclusion}
In this paper, we presented a machine learning approach to model selection for zero-shot cross-lingual transfer, which is appropriate when small amounts of development data are available in one or more pivot languages, but not in the target language.  We showed that our approach improves over the standard practice of model selection using source language development data.  Experiments on five well-studied NLP tasks show that by inspecting internal representations, our method consistently selects better models. LMS also achieves comparable results to the more expensive alternative of annotating small amounts of target-language development data.  Finally, we demonstrated that LMS selects better models for low-resource languages, such as Quechua and Maori, that are not included during pretraining.

\section*{Acknowledgements}
We would like to thank Wei Xu and anonymous reviewers for their valuable suggestions.
This material is based in part on research sponsored by the NSF (IIS-1845670) and IARPA via the BETTER program (2019-19051600004). The views and conclusions contained herein are those of the authors and should not be interpreted as necessarily representing the official policies, either expressed or implied, of IARPA, or the U.S. Government. The U.S. Government is authorized to reproduce and distribute reprints for government purposes notwithstanding any copyright annotation therein.

\bibliographystyle{acl_natbib}
\bibliography{anthology}

\appendix

\clearpage
\newpage
\section{Appendix}
\subsection{Data for Relation Extraction and Argument Role Labeling} 
\label{appendix_imb_data}
In this section, we describe details of the dataset for RE and ARL. Table~\ref{table:stats} reports the statistics of the dataset and Table~\ref{table:imb_gcn} summarizes references and baseline results.

We create a dataset using the ACE2005 corpus \citep{walker2006ace}, which more closely replicates the setting a model will be faced with in a real-world information extraction scenario. 
First, we shuffle documents into 80\%/10\%/10\% splits for train/dev/test, then extract candidate entity-pairs from each document. For RE, the first approach in \citet{ye2019re} is adopted to extract negative instances. 
Negative instances whose entity-type combination has never appeared as a positive example in the training data are filtered out. 
For ARL, we create negative instances by pairing each trigger with every entity in a sentence.  
Details on the two datasets are summarized in Table~\ref{table:stats}. 

As a baseline for the dataset, we reimplement the Graph Convolutional Network (GCN) model of \citet{sub2019rearl} using multilingual embeddings learned by fastText~\citep{bo2017fasttext} on Wikipedia ($\text{GCN}_{\text{ReImp}}$).  Tables~\ref{table:imb_gcn} display F1 for zero-shot cross-lingual transfer.

\begin{table}[h!]
    \centering
    \small
    \begin{tabular}{llrrrr}
    \toprule\\
    Task & Lang & Train & Dev & Test & Pos/Neg\\
    \midrule
    &en & 63177 & 10218 & 6861 & 1:8.9\\
    RE &zh & 57824 & 7096& 8162 & 1:9.4 \\
    &ar & 32984& 4482&4638 & 1:8.9 \\
    \midrule
    &en & 21875 & 3345& 2603& 1:2.6\\
    ARL &zh & 15095 & 2226 & 2017& 1:2.7\\
    &ar & 11587 & 1221 & 1568 & 1:2.9\\
    \bottomrule\\
    \end{tabular}
    \caption{Statistics of the dataset. Number of instances and the total positive/negative ratio.}
    \label{table:stats}
\end{table}

\begin{table}[h!]
\scriptsize
\begin{center}
\begin{tabular}{l cc| cc}
\toprule
\multirow{2}{*}{\bf } & \multicolumn{2}{c}{\bf RE (F$_1$)} & \multicolumn{2}{c}{\bf ARL (F$_1$)} \\ 

  & ar & zh & ar & zh \\
  \midrule
    \cmidrule{1-1} 
  $\text{GCN}_{\text{ReImp}}$ & 39.43 & 32.74 & 16.48 & 23.49\\
    \midrule
  \multicolumn{5}{l}{\emph{Model Selection}}\\
  \midrule
  En-Dev &  36.10 & 67.68 & 44.11 & 60.96 \\
  LMS  & \textbf{39.54} & \textbf{70.75} & \textbf{47.08} & \textbf{62.05}  \\
  \midrule
  100-Target & 34.68 & 68.20 & 44.11 & 62.52 \\
  All-Target  &  41.92 &  69.13 & 47.15 & 63.81 \\
\midrule
\end{tabular}
\caption{F1 scores for relation extraction and argument role labeling on the test set. En-Dev/100-Target/All-Target: model selection based on the highest F1 of English dev set/100 target language dev set examples/target language dev set. Ours: model selection based on the highest scores for the target language: $\argmax_m s(m,l_\text{target})$.
  }
 \label{table:imb_gcn}
\end{center}
\end{table}

\subsection{Variance of Different Meta-train/dev/test Split is Relatively Low}
\label{appendix:model_stat}
In this section, we present a statistical analysis of model selection results for POS and QA between different meta-train/dev/test splits. Table~\ref{table:stat} shows LMS on average improves a point of 0.74 relative to En-Dev and a point of 0.44 relative to Pivot-Dev. We found that the variance in end-task performance between different meta-train/dev/test splits is relatively low.

We train a single LMS with pivot languages in $\{ar, de, es, nl, zh\}$ for POS and $\{ar, de, es, zh\}$ for QA, and test it on all the target languages. All the results are reported with mean and standard deviation with five runs (different meta-train/dev/test splits). A $Z$-test is performed to the differences between LMS/Pivot-Dev and En-Dev. LMS is statistically significantly ($p\text{-value}\leq 0.05$) higher than En-Dev baseline across all languages and two tasks while Pivot-Dev fails in three languages. LMS also obtains a lower standard deviation for the model scores except for Swedish (sv) and Vietnamese (vi).

\begin{table*}[ht!]
\centering
\small
\begin{tabular}{ll|c|cll|ccl}
\toprule
    Task  & Lang & Ref  & En-Dev & Pivot-Dev & LMS            & 100-Target & All-Target  & \# All-Target \\ 
\midrule
\multirow{10}{*}{POS (Acc)}
&bg & 87.4 & 87.22\tiny{$\pm$0.24} & \underline{87.35}\tiny{$\pm$0.37} (es) & \textbf{87.75}\tiny{$\pm$0.14}* & 87.72\tiny{$\pm$0.50} & 88.09\tiny{$\pm$0.17} & 1115 \\
&da & 88.3 & 88.59\tiny{$\pm$0.17} & \textbf{88.81}\tiny{$\pm$0.12} (nl)*& \underline{88.74}\tiny{$\pm$0.14}* & 88.74\tiny{$\pm$0.06} & 89.01\tiny{$\pm$0.18} & 322 \\
&fa & 72.8 &\underline{71.58}\tiny{$\pm$0.28} & 71.55\tiny{$\pm$1.02} (es) & \textbf{73.57}\tiny{$\pm$0.13}* & 73.00\tiny{$\pm$0.76} & 73.80\tiny{$\pm$0.31} & 599 \\
&hu & 83.2 & \underline{82.81}\tiny{$\pm$0.37} & 82.34\tiny{$\pm$0.27} (de) & \textbf{83.28}\tiny{$\pm$0.17}* & 83.14\tiny{$\pm$0.17} & 83.18\tiny{$\pm$0.15} & 179 \\
&it & 84.7 & 84.44\tiny{$\pm$0.45} & \underline{84.68}\tiny{$\pm$0.48} (es)* & \textbf{85.04}\tiny{$\pm$0.19}* & 85.14\tiny{$\pm$0.27} & 85.61\tiny{$\pm$0.21} & 489 \\
&pt & 82.1 & 81.87\tiny{$\pm$0.16} & \underline{82.06}\tiny{$\pm$0.23} (es)* & \textbf{82.18}\tiny{$\pm$0.08}* & 82.14\tiny{$\pm$0.34} & 82.36\tiny{$\pm$0.17} & 271 \\
&ro & 84.7 & 83.84\tiny{$\pm$0.30} & \textbf{85.59}\tiny{$\pm$0.48} (es)* & \underline{84.74}\tiny{$\pm$0.04}* & 84.79\tiny{$\pm$0.47} & 85.39\tiny{$\pm$0.16} & 1191 \\
&sk & 83.6 & 83.49\tiny{$\pm$0.42} & \textbf{84.09}\tiny{$\pm$0.50} (es)* & \underline{83.93}\tiny{$\pm$0.28}* & 84.07\tiny{$\pm$0.58} & 84.90\tiny{$\pm$0.33} & 1060 \\
&sl & 84.2 & \underline{84.26}\tiny{$\pm$0.34} & 84.10\tiny{$\pm$0.57} (es) & \textbf{84.91}\tiny{$\pm$0.24}* & 84.23\tiny{$\pm$0.56} & 85.41\tiny{$\pm$0.61} & 735 \\
&sv & 91.3 & 91.37\tiny{$\pm$0.05} & \textbf{91.63}\tiny{$\pm$0.19} (nl)* & \underline{91.55}\tiny{$\pm$0.21}* & 91.58\tiny{$\pm$0.23} & 91.73\tiny{$\pm$0.05} & 504 \\
\midrule
\multirow{2}{*}{QA (F$_{1}$)}
&hi & 43.8 & 39.93\tiny{$\pm$1.45} & 41.40\tiny{$\pm$1.35}(es)* & 42.09\tiny{$\pm$0.91}* & 40.04\tiny{$\pm$0.81} & 42.56\tiny{$\pm$0.26} & 507\\
&vi & 57.1 & 57.18\tiny{$\pm$0.82} & 57.66\tiny{$\pm$2.03}(ar) & 57.73\tiny{$\pm$1.16}* & 58.83\tiny{$\pm$0.90} & 59.12\tiny{$\pm$0.92} & 511\\
\bottomrule
\end{tabular}
\caption{Model scores (mean $\pm$ sd) selected based on LMS for POS and QA over 5 runs. Bold indicates the best score and underline indicates the second best. $*$ indicates the LMS/Pivot-Dev is statistically significantly ($p\text{-value}\leq 0.05$) higher than En-Dev.}
\label{table:stat}
\end{table*}

\subsection{Does this Approach Generalize to XLM-RoBERTa?}
\label{appendix_xlm}
In Section~\ref{exp:res}, we showed that our approach consistently selects better fine-tuned models than those chosen using English development data. To test the robustness of our approach with a different multilingual pre-trained Transformer, we re-train and evaluate using XLM-RoBERTa-base~\citep{conneau2019unsupervised}, with the same settings used for mBERT in Section~\ref{exp:res} for RE and ARL.

\noindent
{\bf RE} In the left section of Table~\ref{table:xlmr}, our approach selects a model with a higher F1 score compared to En-Dev in Chinese and on par with En-Dev in Arabic. 

\noindent
{\bf ARL} In the right section of Table~\ref{table:xlmr}, our approach selects a model with a higher F1 score compared to En-Dev in Arabic but
performs worse on Chinese (En-Dev outperforms the All-Target).  Overall, our approach appears to be effective when used with XLM-RoBERTa.
\begin{table}[h!]
\scriptsize
\begin{center}
\begin{tabular}{l cc | cc}
\toprule
& \multicolumn{2}{c}{\bf RE (F$_1$)} &
\multicolumn{2}{c}{\bf ARL (F$_1$)}\\
    & ar & zh & ar & zh \\
   \midrule
   $\text{GCN}_{\text{ReImp}}$ & 39.43 & 32.74 & 16.48 & 23.49\\
   mBERT  & 36.10 & 67.68 & 44.11 & 60.96 \\
   \cmidrule{1-5} 
   \multicolumn{3}{l}{\emph{Model Selection}}\\
   \cmidrule{1-5}
    En-Dev & \textbf{40.79} & 64.48 & 50.65 & \textbf{62.73}\\
    LMS & \textbf{40.79} & \textbf{65.11} & \textbf{52.96}  & 61.92 \\
      \cmidrule{1-5} 
   100-Target & 42.33 & 65.38 & 52.90 & 62.12 \\
   All-Target & 44.66 & 65.75 & 53.09 & 62.27 \\
\midrule
\end{tabular}
\end{center}
\caption{XLM-RoBERTa experiment: F1 of relation extraction and argument role labeling. Model selection results are based on XLM-RoBERTa-base models in the meta-test set.}
\label{table:xlmr}
\end{table}

\subsection{Can Multi-task Learning Help?}
\label{appendix_multi_task}
Our setting does not assume access to the labeled data in the target language for a particular task. However, labeled data in the target language may be available for a relevant auxiliary task, which could help the scoring function learn to better estimate whether a model is a good match for the target language.  

To test whether an auxiliary task in the target language might help to select a better model for the target task, we train the LMS on two tasks: RE and ARL.  Gold rankings on the models are then computed for each language using the pivot languages' dev sets. Also, another ``silver'' ranking is computed for each language using the auxiliary task.  The scoring function is then trained to rank mBERT models for both tasks. To differentiate the two tasks, a variant of the scoring function, $s(m,l,t)$, which concatenates a randomly initialized task embedding with the language embedding is adopted. 

In Table~\ref{table:xtask}, our approach selects a model with a higher F1 score for RE. However, multi-task does not benefit ARL but still outperforms En-Dev. As for future direction, we believe an LMS that is trained on an auxiliary dataset can be transferred to the target dataset, hence release the requirement of a small amount of pivot language development data in the target dataset.

\begin{table}[h!]
\scriptsize
\begin{center}
\begin{tabular}{lc|cc}
\toprule
 \emph{Task} & \emph{En-Dev}& ({\tt [Pivot], lang2vec})&  + Multi-task \\
 \midrule
 RE & 51.81 & 55.92 & \textbf{57.31} \\
 ARL & 50.98 & \textbf{53.60} & 51.99  \\
 \midrule
\end{tabular}
\end{center}
\caption{Multi-task analysis using additional training data in the target language from another task. ({\tt [Pivot], lang2vec}): baseline of training within a single task data. Model selection is based on the highest scores for the target language and target task: $\argmax_m s(m,l_\text{target},t_\text{target})$
  }
  \label{table:xtask}
\end{table}

\end{document}